%% file: acl_latex.tex
\documentclass[11pt]{article}

\usepackage[]{acl}

\usepackage{times}
\usepackage{latexsym}

\usepackage{times}
\usepackage{helvet}
\usepackage{courier}
\usepackage{xspace}
\usepackage{natbib}
\usepackage[english]{babel}
\usepackage{amsfonts}
\usepackage{graphicx}
\usepackage{amsfonts,amssymb} 
\usepackage{algorithm2e}
\usepackage{bbm}
\usepackage{color}
\usepackage[inline]{enumitem}
\usepackage{makecell}
\usepackage{tablefootnote}

\usepackage{pifont}
\usepackage{circledsteps}

\usepackage{adjustbox}

\RestyleAlgo{ruled} 
\newcommand*{\eg}{\textit{e.g.}\@\xspace}
\newcommand*{\ie}{\textit{i.e.}\@\xspace}
\newcommand{\method}{\texttt{$\sfrac{2}{3}$-Triplet}\xspace}
\newcommand{\methodext}{\texttt{\sfrac{2}{3}-Triplet+ext}\xspace}
\newcommand{\data}{EMMT\xspace}
\newcommand{\soft}{\textsc{Fusion-based}\xspace}
\newcommand{\hard}{\textsc{Prompt-based}\xspace}

\newcommand{\fusion}{\textsc{Fusion}\xspace}
\newcommand{\prompt}{\textsc{Prompt}\xspace}

\newcommand{\MR}[3]{\multirow{#1}{#2}{#3}}

\newcommand{\T}{\texttt}

\usepackage{graphicx}
\usepackage{multirow}
\usepackage{booktabs}
\usepackage{amsfonts}
\usepackage{verbatim}
\usepackage{pgfplots}
\usepackage{wrapfig}
\usepackage{amssymb,mathrsfs}
\usepackage{array}
\usepackage{pgfplotstable}
\usepackage{pgf}
\usepackage{bm}
\usepackage[normalem]{ulem}
\usepackage{colortbl}
\usepackage{soul}
\usepackage{ulem}
\usepackage{xspace}
\usepackage{xfrac}
\usepackage{pifont}

\usepackage{graphics}

\usepackage{CJKutf8}

\usepackage[caption=false]{subfig}

\usepackage[normalem]{ulem}

\usepackage{threeparttablex}

\usepackage[T1]{fontenc}

\usepackage[utf8]{inputenc}

\usepackage{microtype}

\title{Beyond Triplet: Leveraging the Most Data for Multimodal Machine Translation}

\author{Yaoming Zhu, ~Zewei Sun, ~Shanbo Cheng, ~Luyang Huang, ~Liwei Wu, ~Mingxuan Wang \\ByteDance \\  
\texttt{\{zhuyaoming,sunzewei.v,chengshanbo\}@bytedance.com} \\
\texttt{\{huangluyang,wuliwei.000,wangmingxuan.89\}@bytedance.com}
}
\pgfplotsset{compat=1.18}
\begin{document}
\begin{CJK*}{UTF8}{gkai}

\maketitle
\begin{abstract}
\input{000abstract.tex}
\end{abstract}

\section{Introduction}
\input{010intro.tex}

\section{Related Work}
\input{020related}


\section{Approach}
\input{040approach.tex}

\section{Dataset}
\input{050dataset}

\section{Experiments}
\input{060experiments.tex}

\section{Discussion}

\input{070analysis.tex}

\section{Conclusion}
\input{080conclusion.tex}


\section*{Limitation}
\label{sec:limitation}
\input{085limitation.tex}

\bibliography{custom}
\bibliographystyle{acl_natbib}

\cleardoublepage

\section*{Appendix}
\input{090appendix.tex}
\end{CJK*}

\end{document}

%% file: 000abstract.tex
Recent work has questioned the necessity of visual information in Multimodal Machine Translation (MMT).   
This paper tries to answer this question and build a new benchmark in this work. 
As the available dataset is simple and the text input is self-sufficient, we introduce a challenging dataset called \data, whose testset is deliberately designed to ensure ambiguity.
More importantly, we study this problem in a real-word scenario towards making the most of multimodal training data. We propose a new framework \method which can naturally make full use of large-scale image-text and parallel text-only data.
Extensive experiments show that visual information is highly crucial in \data. The proposed \method outperforms the strong text-only competitor by 3.8 BLEU score, and even bypasses a commercial translation system. 
\footnote{Codes and data are available at \url{github.com/Yaoming95/23Triplet} and \url{huggingface.co/datasets/Yaoming95/EMMT} }


%% file: 010intro.tex
Multimodal Machine Translation (MMT) is a  machine translation task that utilizes data from other modalities, such as images.
Previous studies propose various methods to improve translation quality by incorporating visual information and showing promising results~\citep{LinMSYYGZL20,CaglayanKAMEES21,LiLZZXMZ22,JiaYXCPPLSLD21}. 
However, manual image annotation is relatively expensive; at this stage, most MMT work is applied on a small and specific dataset, Multi30K~\cite{ElliottFSS16}.  The current performance of the MMT system still lags behind the large-scale text-only Neural Machine Translation (NMT) system, which hinders the real-world applicability of MMT.

\begin{figure}
	\centering
	\includegraphics[width=0.9\linewidth]{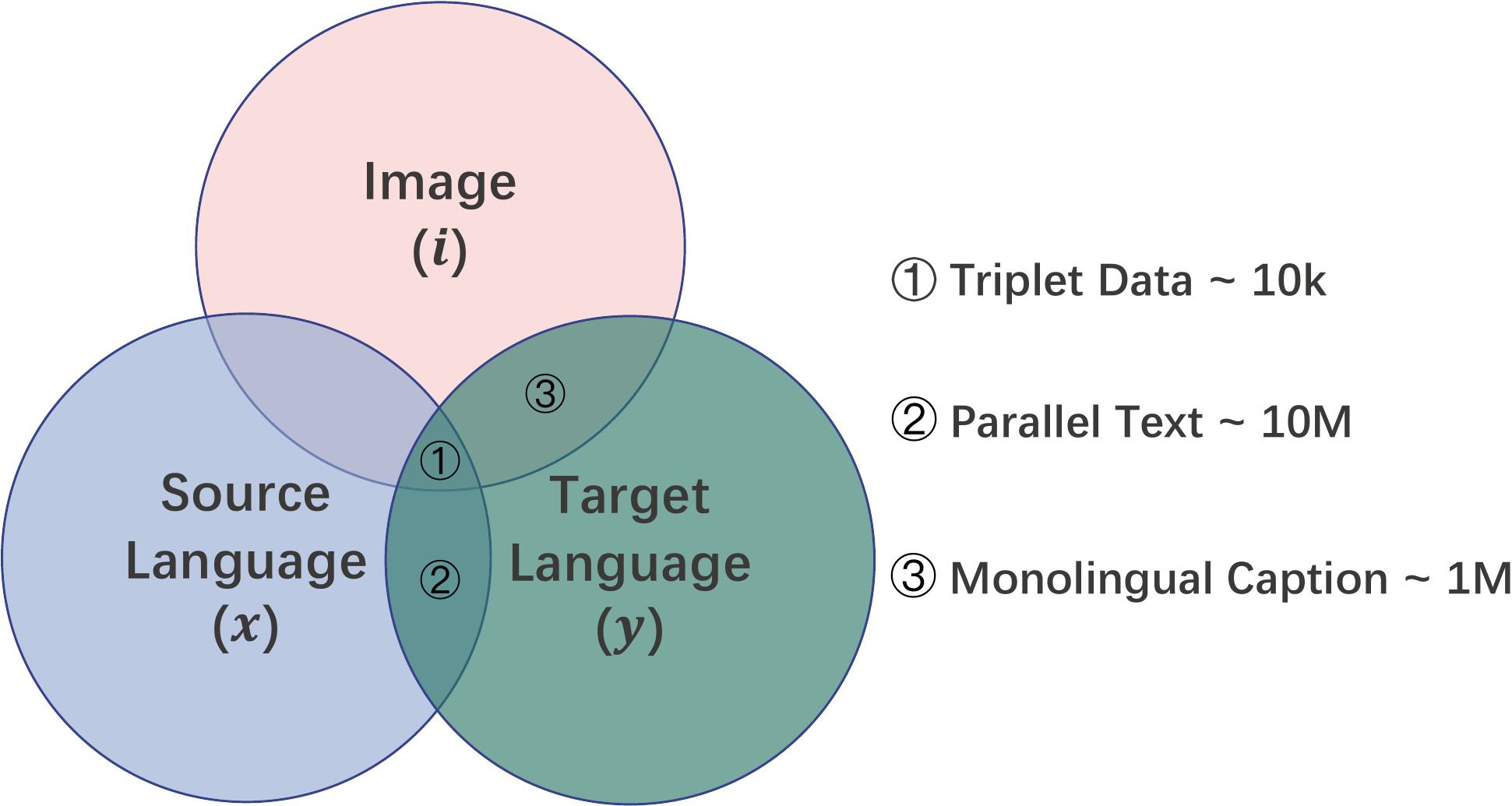}
	\captionsetup{font={footnotesize}}
	\caption{Triple data, although widely utilized in multimodal machine translation,  is quite scarce. We emphasize the importance of other two kinds of data: parallel text and image captions. The numbers represent the size of commonly used datasets for the corresponding data type.  }
\label{fig:intro}
\end{figure}

We summarize the limitations of the current MMT in two aspects.
The first limitation is the size of the training data. Usually, the performance of MMT heavily relies on the triple training data: parallel text data with corresponding images. The triplets are much rarer for collection and much more costly for annotation than monolingual image-text and parallel text data, as in Figure~\ref{fig:intro}. Considering that current MT systems are driven by a massive amount of data~\citep{AharoniJF19}, the sparsity of multimodal data hinders the large-scale application of these systems. Some researchers have proposed retrieve-based approaches~\citep{0001C0USLZ20,FangF22}, aiming to construct pseudo-multimodal data through text retrieval. However, their constructed pseudo-data face problems like visual-textual mismatches and sparse retrieval. Besides, the models still cannot take advantage of monolingual image-text pairs.

The second limitation is the shortage of proper benchmarks. Although several researchers have examined the benefit of visual context upon the translation when textural information is degradated~\cite {CaglayanMSB19,WangX21}, the improvements remain questionable. \citet{WuKBLK20} and \citet{LiAS21} argue that vision contributes minor in previous MMT systems, and the images in the previous benchmark dataset provide limited additional information. In many cases, the translation of sentences relies on textual other than image information. The texts contain complete contexts and are unambiguous, leaving the usage of images doubtful. 
Therefore, a benchmark in that the sentences can not be easily translated without visual information is much needed.

To address these limitations, we propose models to make the most of training data and build a challenge and real-world benchmark to push the real-world application of MMT research.  
At first, we propose a new framework, named \method, which can use both parallel text and image-text data. 
It provides two different ways of exploiting these data based on the continuous vision feature and discrete prompt tokens, respectively. The two approaches are not mutually exclusive and can be used jointly to improve performance within the same framework. It is also worth mentioning that the prompt approach is easy to deploy without modifying the model architecture. 

In addition,  we present a new real-world dataset named \data. We collect parallel text-image data from several publicly available e-commerce websites and label the translation by $20$ language experts. 
To build a challenge test set,  we carefully select ambiguous sentences that can not be easily translated without images. 
This high-quality dataset contains 22K triplets for training and 1000 test examples, along with extra image-text and parallel text data.


Comprehensive experiments show that \method rivals or surpasses text-only and other MMT competitors on \data, as well as previous benchmarks.  
Especially, \method consistently improves the strong text-only baseline by more than 3 BLEU scores in various settings, showing the importance of visual information.

%% file: 020related.tex


Researchers applied multimodal information to enhance machine translation systems since the statistical machine translation era~\citep{HitschlerSR16,AfliBS16}. With the rise of neural networks in machine translation, researchers have focused on utilizing image information more effectively. Early work used image features as initialization for neural MT systems~\citep{LibovickyH17}. More recent studies proposed multimodal attention mechanisms~\citep{CalixtoLC17,YaoW20}, enhanced text-image representations using graph neural networks~\citep{LinMSYYGZL20}, latent variable models or capsule networks~\citep{YinMSZYZL20}, and used object-level visual grounding information to align text and image~\citep{WangX21}. 
\citet{LiLZZXMZ22} found that a stronger vision model is more important than a complex architecture for multimodal translation. 

As we discussed earlier, these methods are limited to bilingual captions with image data, which is scarce. Therefore, some researchers~\citep{0001C0USLZ20,FangF22} also design retrieval-based MMT methods that retrieve images with similar topics for image-free sentences. 
Alternatively, \citet{ElliottK17} proposed visual ``imagination'' by sharing visual and textual encoders. 

Recently, \citet{WuKBLK20} and \citet{LiAS21} have questioned whether the most common benchmark Multi30K~\citep{ElliottFSS16} is suited for multimodal translation since they found images contribute little to translation.
\citet{0003CJLXH21} have contributed a new dataset of the e-commercial product domain. However, we find their datasets still have similar drawbacks. 

Several relevant studies about translation and multimodality are noteworthy.
\citet{HuangHCH20} used visual content as a pivot to improve unsupervised MT. 
\citet{wang2022simvlm} proposed a pre-training model by using modality embedding as prefix for weak supervision tasks.
\citet{li2022valhalla} introduced the VALHALLA, which  translates under guidance of hallucinated visual representation.

%% file: 040approach.tex


\begin{figure*}[t]
\centering

\subfloat[\soft \label{fig:fusion}]
{%
  \includegraphics[width=0.47\linewidth]{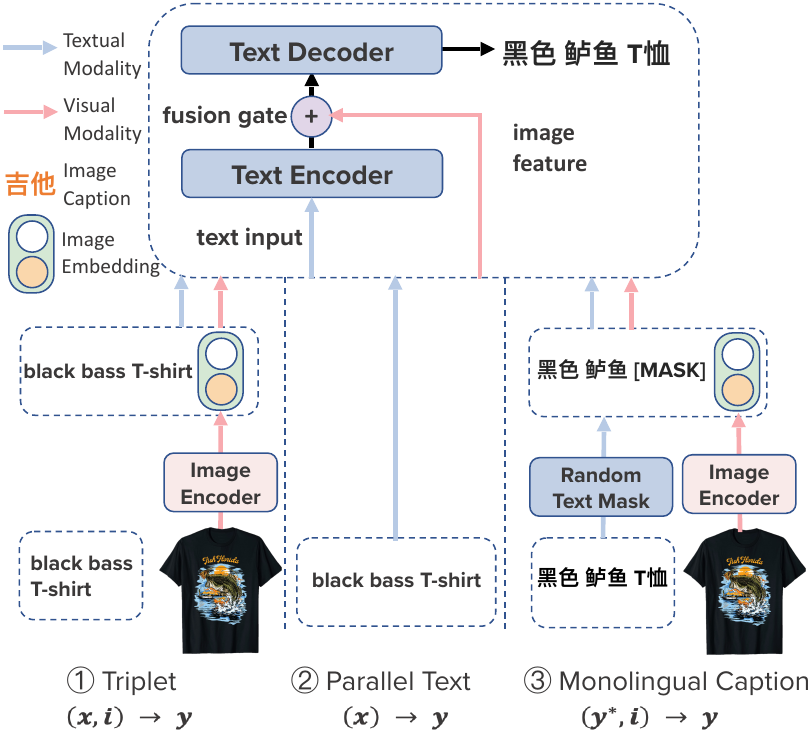}
} \qquad
\subfloat[\hard \label{fig:prompt}]
{%
  \includegraphics[width=0.47\linewidth]{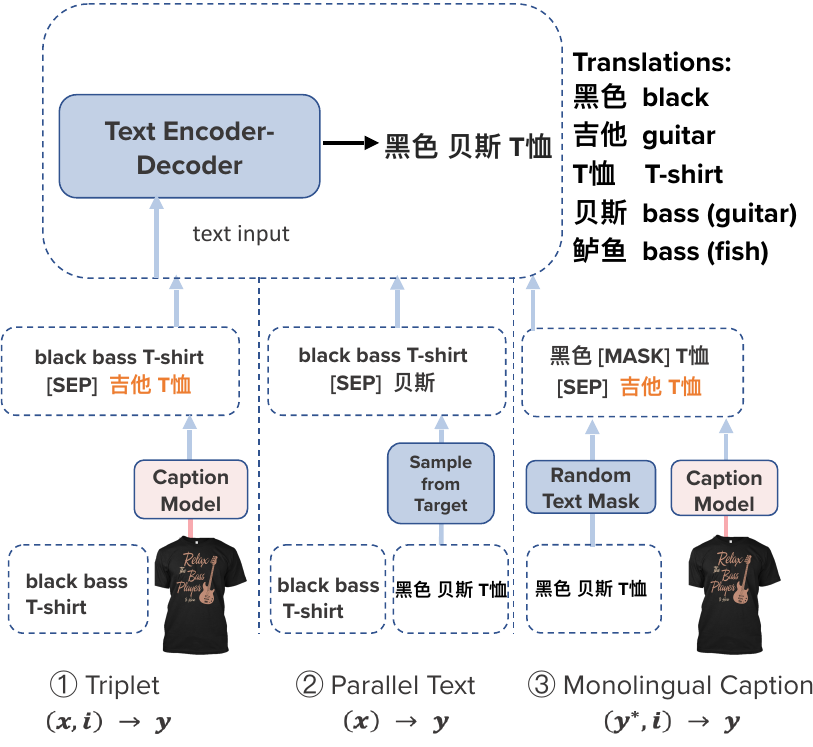}%
}
\caption{ The illustration of our framework \method in \soft and \hard given ambiguous sentences. For each approach, the model can conduct a mixed training with three kinds of data: $\{(x, y, i)\}$, $\{(y, i)\}$, and $\{(x, y)\}$. \ding{172} means using triple data ($(x, i)\rightarrow y$); \ding{173} means using parallel text data ($(x)\rightarrow y$); \ding{174} means using monolingual image-text data ($(y^*, i)\rightarrow y$), where $y^*$ indicates the masked target text.}
\label{fig:framework}
\end{figure*}



For the fully supervised condition in MMT, we have triplet $\{(x, y, i)\}$, where $x$ is the source text, $y$ is the target text, and $i$ is the associated image. Since the triplet is rare, we attempt to utilize partially parallel data like $\{(y, i)\}$ and $\{(x, y)\}$, which are referred as monolingual image-text data and parallel text data in this paper.

In this section, we propose a new training framework \method with two approaches to utilize triple and non-triple data at the same time.
We name these two approaches as \textbf{\soft} and \textbf{\hard}, as shown in Figure~\ref{fig:framework}. 

For each approach, the model can conduct a mix training with three kinds of data: ($(x, i)\rightarrow y$), ($(x)\rightarrow y$), and ($(y^*, i)\rightarrow y$), where $y^*$ indicates the masked target text.

\soft approach resembles the conventional models where the encoded vision information is taken as model input and the model is trained in end2end manners, and our design makes it possible to utilize bilingual corpus and image-text pairs other than multilingual triplets. 

\hard approach is inspired by the recent NLP research based on prompts~\citep{GaoFC20,LiL20,WangSCZW22,SunJHCCW22}, where we directly use the image caption as a prompt to enhance the translation model without any modification to the model.





\subsection{\soft}
The common practice to utilize image information is to extract vision features and use them as inputs of the multimodal MT systems.  
Typically, it's common to cooperate vision and textual features to get a multimodal fused representation, where the textual features are the output state from the Transformer encoder and the vision feature is extracted via a pre-trained vision model.

We incorporate textual embedding and image features by simple concatenation:

\begin{equation}\label{fuse}
H^{\mathrm{fused}} = [H^{\mathrm{text}} ; \textbf{h}^{\mathrm{img} } ]
\end{equation}

where $ H^{\mathrm{text}}$ is the encoded textual features of Transformer encoder, and $\textbf{h}^{\mathrm{img} }$ is the visual representation of [CLS] token broadcated to the length of the text sequence. 

Then, we employ a gate matrix $ \Lambda $ to regulate the blend of visual and textual information. 

\begin{equation}\label{gate}
 \Lambda = \tanh(f( [H^{\mathrm{text}} ; H^{\mathrm{fused}}  ]))
\end{equation}

Finally, we add the gated fused information to the origin textual feature to get the final multimodal fused representation:

\begin{equation}\label{main}
H^{\mathrm{out}} = H^{\mathrm{text}} + \mathbbm{1}(\mathrm{img})  \Lambda H^{\mathrm{fused}}
\end{equation}

$\mathbbm{1}(\mathrm{img}) $ indicates whether the image exists. The value is set to zero when image is absent. 

It is worth noting that in Eq.\ref{gate}, we employ the hyperbolic tangent (tanh) gate instead of the traditional sigmoid gate~\citep{WuKBLK20,LiLZZXMZ22} in the multimodal translation scenario. 
The new choice has two major advantages: (a) The output of the tanh can take on both positive and negative values, thereby enabling model to modulate the fused features $H^{\mathrm{fused}}$ in accordance with the text $H^{\mathrm{text}}$; (b) The tanh function is centered at zero, thus, when the fused feature is close to zero, the output of the gate is also minimal, which aligns with the scenario where the image is absent naturally (\ie $\tanh(0) = \mathbbm{1}(\mathrm{no\ img})=0$).

The next paragraphs illustrate how to utilize three types of data respectively.

\paragraph{Using Triple Data ($(x, i)\rightarrow y$)}~Figure~\ref{fig:fusion} \Circled{1}: Based on the basic architecture, we take in the source text for the text encoder and the image for the image encoder. By setting $\mathbbm{1}(\mathrm{img})=1$, we naturally leverage vision context for translation. The inference procedure also follows this flow. 

\paragraph{Using Parallel Text ($(x)\rightarrow y$)}~
Figure~\ref{fig:fusion} \Circled{2}:
We utilize the same architecture as the triple data setting. By setting $\mathbbm{1}(\mathrm{img})=0$, we can adapt to the text-only condition. For the image-free bilingual data, the fused term is absent, and the final representation $H^{\mathrm{out}} $ is reduced to textual only, consistent with the learning on unimodal corpus. 

\paragraph{Using Monolingual Caption ($(y^*, i)\rightarrow y$)}
Figure~\ref{fig:fusion} \Circled{3}:
Inspired by \citet{SiddhantBCFCKAW20}'s strategy on leveraging monolingual data for translation, we adapt the mask de-noising task for utilizing monolingual image-text pairs. In a nutshell, we randomly mask some tokens in the text, and force the model to predict the complete caption text based on the masked text and image as input.

\subsection{\hard}

As prompt-based methods have made great success in NLP tasks\citep{GaoFC20,LiYL022,WangSCZW22,SunJHCCW22},  we also consider whether the image information can be converted to some prompt signals for guiding sentence generation. 

The general idea is quite straight: our translation system accepts a sentence of source language along with some keywords of target language, and translates the source sentence into the target language under the instruction of the target keywords. The keywords can be any description of the image that can help disambiguate the translation.

\paragraph{Using Triple Data ($(x, i)\rightarrow y$)}~Figure~\ref{fig:prompt} \Circled{1}:
First, we generate the prompt from the image with a pre-trained caption model (we will introduce the caption model later). The source sentence is concatenated with the  
The original source sentence and the prompt are concatenated together to compose the training sources, with a special token \texttt{[SEP]} as a separator between the two. 

\paragraph{Using Parallel Text ($(x)\rightarrow y$)}~
Figure~\ref{fig:prompt} \Circled{2}:
Since \hard approach adopts a standard Transformer and involves no modification on architecture, it is natural to train on unimodal parallel corpus. We use the parallel data to strengthen the ability to take advantage of the prompt. Without any image, we randomly select several words from the target sentence as the pseudo vision prompt. For translation training, we append the keyword prompt to the end of the original sentence and use a special token as a separator~\citep{LiYL022}. After inference, we extract the translation result by splitting the separator token.

\paragraph{Using Monolingual Caption ($(y^*, i)\rightarrow y$)}~
Figure~\ref{fig:prompt} \Circled{3}:
Like \soft approach, we use the de-noising auto-encoder task. By randomly masking some tokens and combining the caption result from the image as the prompt, we make the model learn to predict the original target text.

\paragraph{Training Caption Model ($(i)\rightarrow \text{keywords}(y)$)}
Meanwhile, we train an caption model to generate the guiding prompt from images for translation, 
We formulate image-text pairs from both triple data and target-side monolingual caption. 
The input and output of the model are the image and extracted keywords of the corresponding target sentence. 


\subsection{Comparison and Combination of \soft and \hard}

Under the same training framework \method, we propose two approaches, {\soft} and {\hard}, for utilizing non-triple data. The {\soft} approach preserves the complete visual context, providing more information via model fusion. In contrast, the {\hard} approach has the advantage of not requiring any modifications to the model architecture. Instead, all visual information is introduced by the prompt model, making deployment more straightforward.

The two methods, {\soft} and {\hard}, are not mutually exclusive, and we can jointly utilize them. Specifically, the model simultaneously utilizes the fused feature in Eq.~\ref{main} as an encoder representation and the prompted-concatenated source as text input. The combination enables the model to benefit from our framework in the most comprehensive way, and as a result, the performance gains significant improvements.




%% file: 050dataset.tex
As mentioned before, in previous test sets, many sentences can be easily translated without the image context, for all information is conveyed in the text and has no ambiguity. To deeply evaluate visual information usage, we propose a multimodal-specific dataset.

We collect the data based on real-world e-commercial data crawled from TikTok Shop and Shoppee. We crawled the product image and title on two websites, where the title may be in English or Chinese. 
We filter out redundant, duplicate samples and those with serious syntax errors.
Based on this, we conduct manual annotations. We hired a team of 20 professional translators. All translators are native Chinese, majoring in English. 
In addition, another translator independently samples the annotated corpus for quality control. 
We let the annotators select some samples specifically for the test set, which they found difﬁcult to translate or had some confusion without images. 
The total number of triples annotated is 22, 500 of which are carefully selected samples as testset. We also randomly selected 500 samples as devsets among the full-set while the remaining as training set. 

Besides the annotated triplets, we clean the rest of the crawled data and open sourced it as the monolingual caption part of the data.
Since our approach features in utilizing bilingual data to enhance multimodal translation, we sample 750K
 CCAlign~\citep{El-KishkyCGK20} English-Chinese as a bilingual parallel text. The selection is motivated by the corpus's properties of its diversity in sources and domains, and it is more relevance to real-world compared to other corpus.
The sampled data scale is decided based on both the model architecture and the principles of the neural scaling law~\citep{abs-2001-08361,GordonDK21}. We also encourage future researchers to explore the use of additional non-triple data to further enhance performance, as detailed in the appendix.
We summarize the dataset statistics in Table~\ref{Tab:statistics}. We discuss ethic and copyright issue of the data in the appendix.

%% file: 060experiments.tex
\subsection{Datasets}
We conduct experiments on three benchmark datasets: Multi30K~\citep{ElliottFSS16}, Fashion-MMT (Clean)~\citep{0003CJLXH21}, and our \data.
\textbf{Multi30K} is the most common benchmark on MMT tasks, annotated from \textit{Flickr}, where we focus on English-German translation. To validate the effectiveness of parallel text, we add 1M English-German from CCAlign and COCO~\cite{LinMBHPRDZ14,BiswasBHS21}.
\textbf{Fashion-MMT} is built on fashion captions of FACAD~\citep{YangZJLWTXWW20}.

\subsection{Baselines}
We compare our proposed \method with the following SOTA MT and MMT systems:  

\noindent     \textbf{Transformer}~\citep{VaswaniSPUJGKP17} is the current \textit{de facto} standard for text-based MT.
    
\noindent     \textbf{UPOC$^2$}~\citep{0003CJLXH21} introduced cross-modal pre-training tasks for multimodal  translation.
    
\noindent     \textbf{Selective-Attention} (SA)~\citep{LiLZZXMZ22} investigated strong vision models and enhanced features can enhance multimodal translation with simple attention mechanism.
    
\noindent    \textbf{ UVR-NMT}~\citep{0001C0USLZ20} retrieves related images from caption corpus as the pseudo image for sentences.
    
\noindent     \textbf{Phrase Retrieval}~\citep{FangF22} is an improved version of retrieval-based MMT model that retrieve images in phrase-level.

In addition, we report the results of Google Translate, which helps to check whether the translation of the test set actually requires images. All baselines reported use the same number of layers, hidden units and vocabulary as \method for fair comparison. 

We mainly refer to BLEU~\citep{PapineniRWZ02} as the major metric since it is the most commonly used evaluation standard in various previous multimodal MT studies.

\input{tables/stat.tex}

\input{tables/main3-v2.tex}

\subsection{Setups}
\label{sec:setups}

To compare with previous SOTAs, we use different model scales on Multi30K and the other two datasets. We follow \citet{LiLZZXMZ22}'s and \citet{LiAS21}'s setting on Multi30K, where the model has 4 encoder layers, 4 decoder layers, 4 attention heads, hidden size and filter size is 128 and 256, respectively. On the other two datasets, we set the model has 6 encoder layers, 6 decoder layers, 8 attention heads, hidden size and filter size is 256 and 512, respectively~(\ie Transformer-base setting). We apply BPE~\citep{SennrichHB16a} on tokenized English and Chinese sentences jointly to get vocabularies with 11k merge operations.
We use \citet{ZengZL22}'s method to get the caption model.  
The vocabularies, tokenized sentence and caption models will be released for reproduction. Codes are based on Fairseq~\citep{OttEBFGNGA19}.

When training models on various domains (+PT and +MC in Tab.~\ref{tab:main}), we upsample small-scale data (\ie E-commercial Triplet) because of the massive disparity of data scale in different domains. We follow ~\citet{WangN19}'s and ~\citet{abs-1907-05019}'s temperature based data sampling strategy and set the sampling temperature at 5. 
We empirically find that the model gains by simply randomly dropping some images during the training, where we set the drop ratio at 0.3~. 
Interestingly, such the method is also observed in other multimodal research topics~\citep{AbdelazizTDKAK20,Alfasly_2022_CVPR}. We evaluate the performance with tokenized BLEU~\citep{PapineniRWZ02}.

\subsection{Main Results}

We list the main results in Tab.~\ref{tab:main}. 
We get three major findings throughout the results: 
\begin{enumerate}[itemsep=-2pt]
  \item In traditional multimodal MT settings (\ie Triplet only and Multi30K), whose training and inference are on triple data, \method rivals or even surpasses the previous SOTAs. 
  \item Parallel text and monolingual captions significantly boost the performance of multimodal translation models.
  With these additional data, even the plain Transformer model outperforms SOTA multimodal baselines. Given the scarcity of multimodal data, we argue that the use of extra data, especially the parallel text, is more crucial for multimodal translation than the use of multimodal information.
  \item \fusion and \prompt generally achieve the best performance when used together. This suggests two approaches are complementary.

  
\end{enumerate}

We also list results on Multi30k for dataset comparison. Google Translate achieves the best results, while all other models are close in performance with \textbf{no} statistical significant improvement.
It indicates that images in Multi30K are less essential and a strong text translation model is sufficient to handle the majority of cases. Moreover, we find that by incorporating non-imaged parallel text, the model's performance improves significantly, while narrows the gap between plain transformer models  MMT ones. Hence, the parallel text  rather than images may be more essential for improving performance on the Multi30k. 
In contrast, \method surpass Google's on \data with visual infomation, providing evidence that ours serves as a suitable benchmark.

We also report the results of \method and baselines on FashionMMT in Appendix along with BLEURT~\citep{SellamDP20} and word accuracy as supplementary metrics.
The results show that \method also rivals the SOTA MMT systems on various benchmarks and metrics.

\subsection{Performance on Triplet-unavailable Setting}

In more scenarios, annotated triple data is rather scarce or even unavailable, \ie only bilingual translation or monolingual image caption is available in the training data, while we wish the model can still translate sentences in multimodal manners.

\begin{figure}[h]
    \centering
    \includegraphics[width=0.91\linewidth]{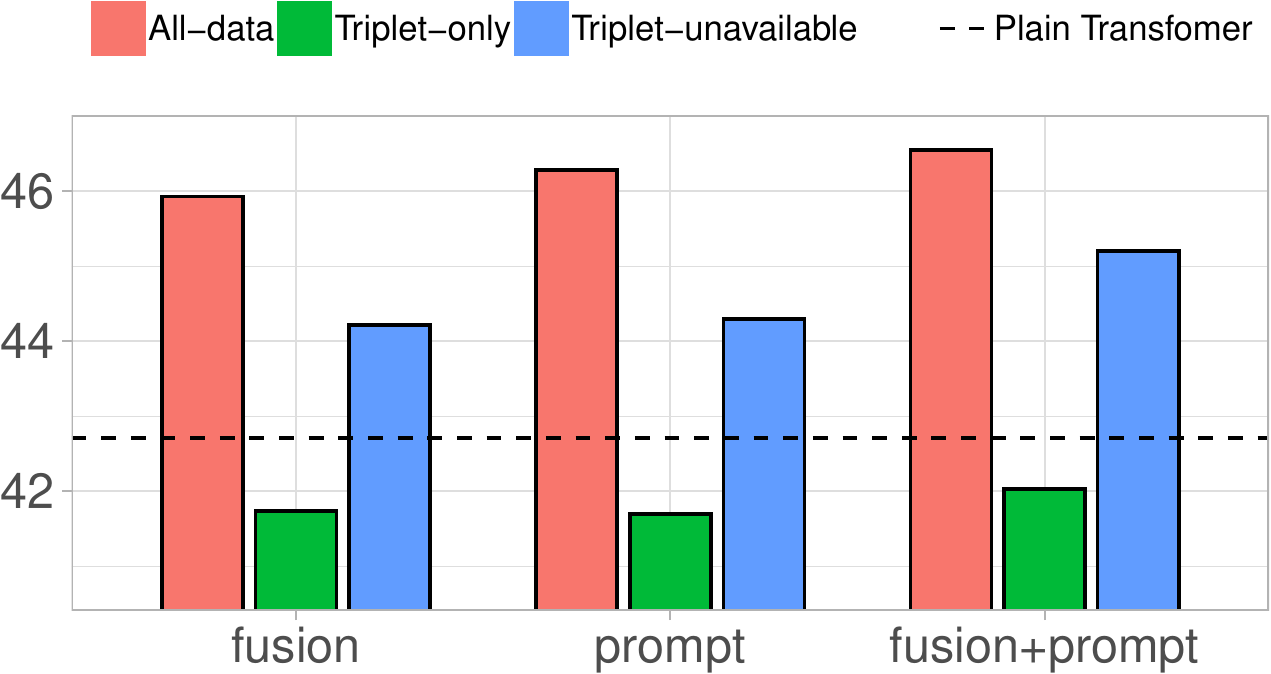}
    
    \caption{\method's performance on using all-data, use triplet-only and triplet-unavailable cases. The horizontal line is the transformer baseline trained on all-data(\ie +PT+MC of Row 1 in Table~\ref{tab:main}) }
    \label{fig:zero}
\end{figure}

Since our proposed \method utilize not only triplets,
we examine whether our model can conduct inference on multimodal triple testset while only trained on the non-triple data, as triplet might be unavailable in real scenarios. 
In this experiment, we discard all images of \data's triples during the training stage, while the trained model is still evaluated on the multimodal test set. We compare the triplet-unavailable results to triplet only and full data training set settings in Figure~\ref{fig:zero}

We can see that \method still preserves a relatively high performance and even sharply beats the triplet-only setting. This fully illustrates that involving parallel text and monolingual caption is extremely important for MMT.

%% file: tables/stat.tex
\begin{table}[]
\centering
\begin{tabular}{ccc|cc}
\Xhline{3\arrayrulewidth}
\multicolumn{3}{c|}{Train}       & \multirow{2}{*}{Test} & \multirow{2}{*}{Dev} \\ \cline{1-3}
Triplet & PT & MC &      &     \\ \hline
22K   & 750K     & 103K        & 1000  & 500 \\ 
\Xhline{3\arrayrulewidth}
\end{tabular}
\caption{\data statistics. ``PT'' stands for parallel text data. ``MC'' stands for monolingual caption.}
\label{Tab:statistics}
\end{table}

%% file: tables/main3-v2.tex
\begin{table*}[]
\centering
\begin{adjustbox}{max width=0.9\textwidth}
\begin{threeparttable}[b]
\begin{tabular}{c|l||ccc||cc|cc}
\Xhline{3\arrayrulewidth}
\multirow{2}{*}{ID} & Test set            & \multicolumn{3}{c||}{EMMT}                        & \multicolumn{2}{c}{Multi30k-Test16} & \multicolumn{2}{c}{Multi30k-Test17}               \\ \cline{2-9} 
                    & Training Data       & Triplet Only   & +PT            & + PT + MC      & Triplet Only    & +PT      & Triplet Only & +PT   \\ \hline
1                   & Plain Transformer\tnote{$\heartsuit$}   & 39.07          & 40.66          & 42.71          & 39.97           & 44.13    & 31.87        & 40.46 \\
2                   & Selective Attention\tnote{$\spadesuit$} & 41.27          & /              & /              & 40.63           & /        & 33.80        & /                         \\
3                   & UPOC$^2$\tnote{$\diamondsuit$}            & 40.60          & /              & 44.81              & 40.8            & /        & 34.1         & /                         \\ \hline
4                   & UVR-NMT\tnote{$\clubsuit$}             & 37.82          & 41.13         & /              & 38.19           & /        & 31.85        & /                         \\
5                   & Phrase Retrieval\tnote{$\clubsuit$}    & /              & /              & /              & 40.30           & /        & 33.45        & /                         \\ \hline
6                   & \soft                & 41.74          & 44.22          & 45.93          & 40.95           & /        & 34.03        & /                         \\
7                   & \hard                & 41.70          & 43.35          & 46.28          & 40.17           & /        & 33.87        & /                         \\
8                   & \textsc{Fusion+Prompt}           & \textbf{42.03} & \textbf{45.20} & \textbf{46.55} & 40.48           & 44.60    & 34.62        & 40.07 \\ \hline
                    & Google Translate    & \multicolumn{3}{c||}{44.27}                       & \multicolumn{2}{c|}{41.9}   & \multicolumn{2}{c}{42.0}                 \\ \hline
\Xhline{3\arrayrulewidth}
\end{tabular}
 \begin{tablenotes}
 \scriptsize
 \item [$\heartsuit$] We also train plain Transformer on monolingual captions via \citet{SiddhantBCFCKAW20}'s method for fair comparison on textual data. 
 \item [$\spadesuit$] We use their open source code to reproduce Multi30K's results.
\item [$\diamondsuit$] Multi30K's results copy from ~\citet{0003CJLXH21}. We add all MC and PT data for its pre-training in +PT+MC column for fair comparison on data. The complete UPOC$^2$ also utilize product attributes besides images, which is removed from our replication.
 \item [$\clubsuit$] Multi30K's results copy from ~\citet{FangF22}. Phrase Retrieval is not reported on \data since they haven't released the phrase extraction scripts. We conduct the retrieval for all parallel sentences with top 5 images as candidate in +PT column of UVR-NMT.
 \end{tablenotes}

\caption{Results of \method and related work on \data and Multi30k. ``PT'' indicates parallel text data $\{(x, y)\}$, ``MC'' indicates monolingual caption data $\{(y, i)\}$. On the one hand, \method outperforms previous studies. On the other hand, extra non-triple data brings significant improvements. The reported improvement on \textbf{\data} dataset is examined with \citet{ReichartDBS18}'s significance test with $p<0.05$~. 
}
\label{tab:main}

\end{threeparttable}

 \end{adjustbox}
 
\end{table*}

%% file: 070analysis.tex
As plenty of previous studies have discussed, the current multimodal MT benchmarks are biased, hence the quality gains of previous work might not actually derive from image information, but from a better training schema or regularization effect~\citep{DodgeGCSS19,HesselL20}. 
This section gives a comprehensive analysis and sanity check on our proposed \method and \data: we carefully examine whether and how our model utilize images, and whether the testset of \data has sufficient reliability.

\subsection{Visual Ablation Study: Images Matter}

We first conduct ablation studies on images to determine how multimodal information contributes to the performance. 
Most studies used \textbf{adversarial} input (\eg shuffled images) to inspect importance of visual information. However, effects of adversarial input might be opaque~\citep{LiAS21}.
Hence, we also introduce \textbf{absent} input to examine whether \method can handle source sentences without image by simply zeroing the image feature for \fusion or striping the prompt for \prompt. 

\begin{figure}[t]
\centering
\includegraphics[width=0.91\linewidth]{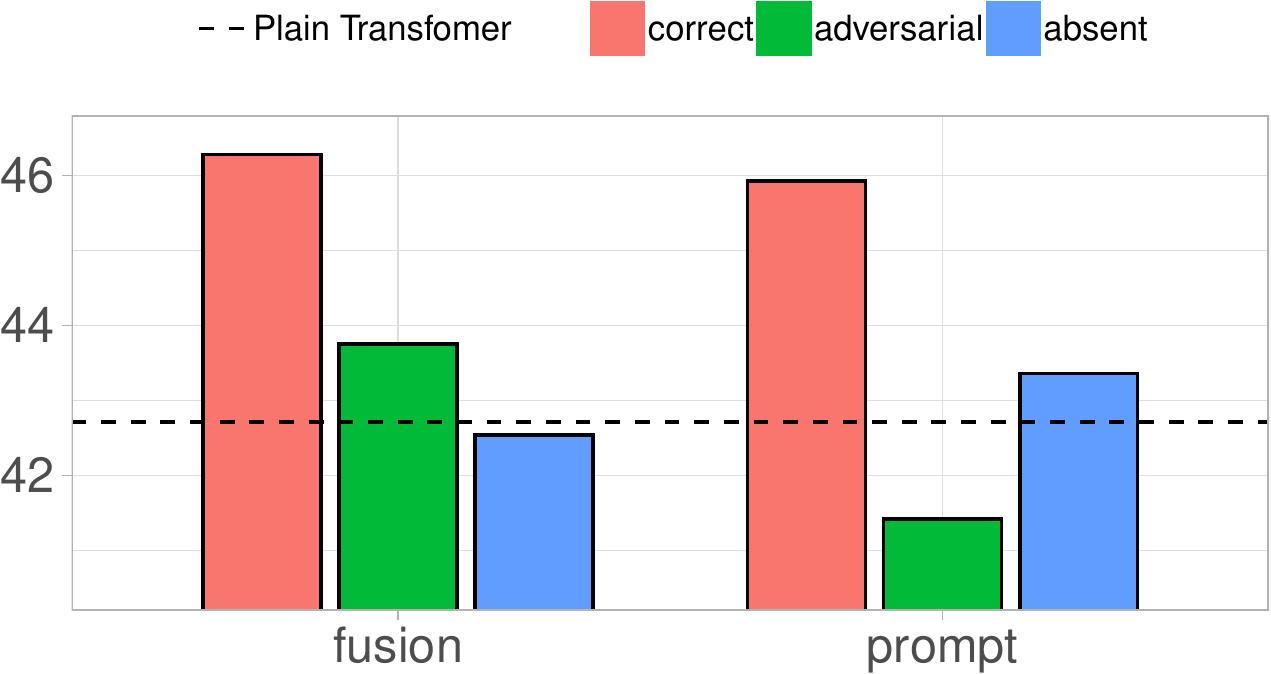}

\caption{The results of ablation study by given empty image (absent) and wrong image (adversarial) as input. }
\label{fig:ablation}
\end{figure}

We list the results of a vision ablation study of both adversarial and absent respectively in Figure~\ref{fig:ablation}, where we select \soft and \hard approaches trained with full data(last columns in Table~\ref{tab:main}) for comparison.

\begin{figure}[h]
\centering
\includegraphics[clip,width=.9\columnwidth]{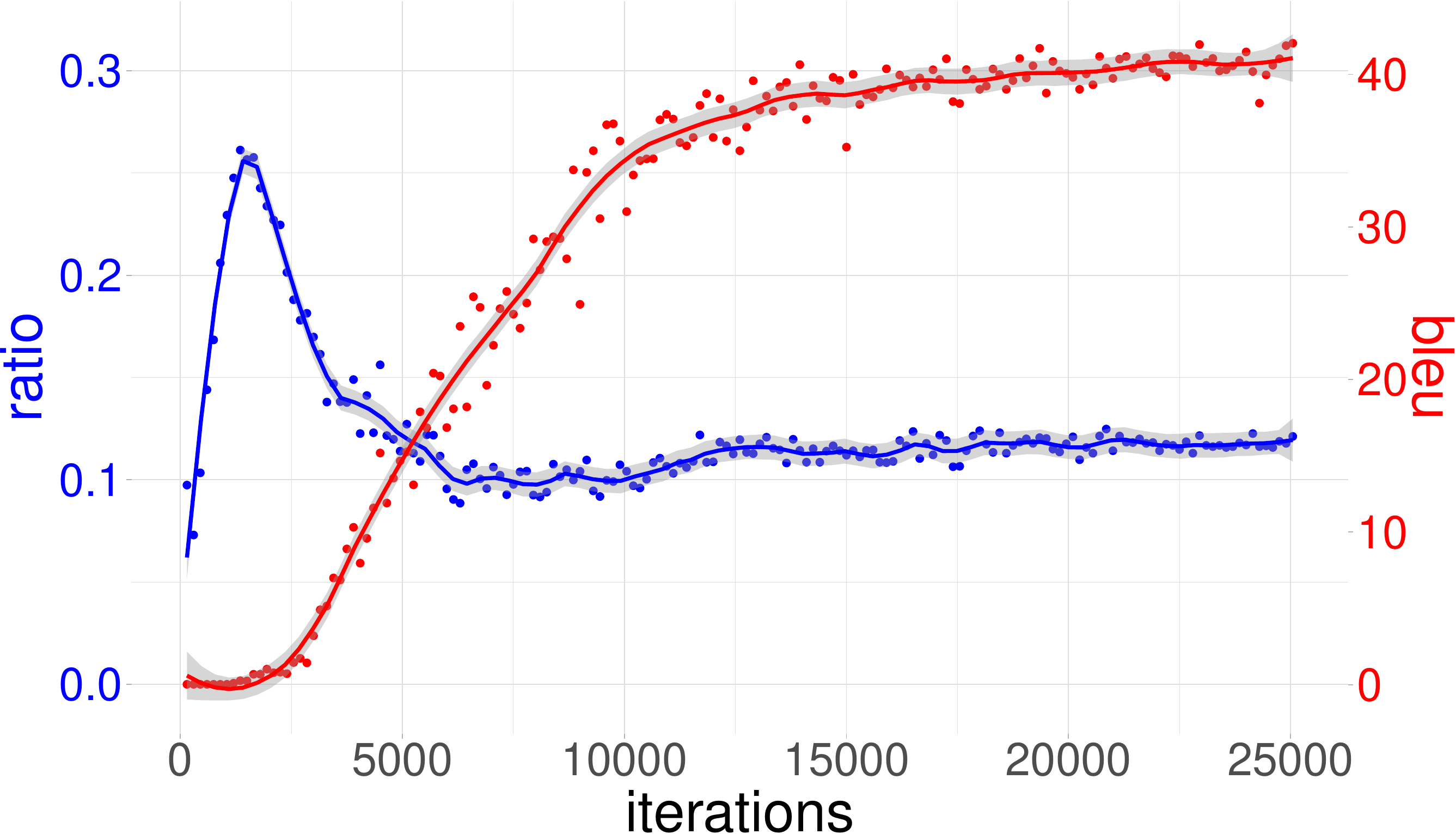}%

\caption{Ratio-BLEU on testset during the training  }
\label{fig:ratio-bleu}

\end{figure}

In the absent setting, both the \fusion and \prompt degrade to the baseline, confirming the reliance of \method on image information. In the adversarial setting, the \prompt performs worse than the baseline, which is in line with the expectation that incorrect visual contexts lead to poor results. However, while the \fusion also exhibits a decline in performance, it still surpasses the baseline. This aligns with the observations made by \citet{Elliott18,WuKBLK20} that the visual signal not only provides multimodal information, but also acts as a regularization term. We will further discuss this issue in Section~\ref{sec:limitation}.

\vspace{-8pt}

\subsection{How Visual Modality Works}

We further investigate how the visual signal influence the model. 
\begin{figure}[]
    \centering
    \includegraphics[width=0.9\linewidth]{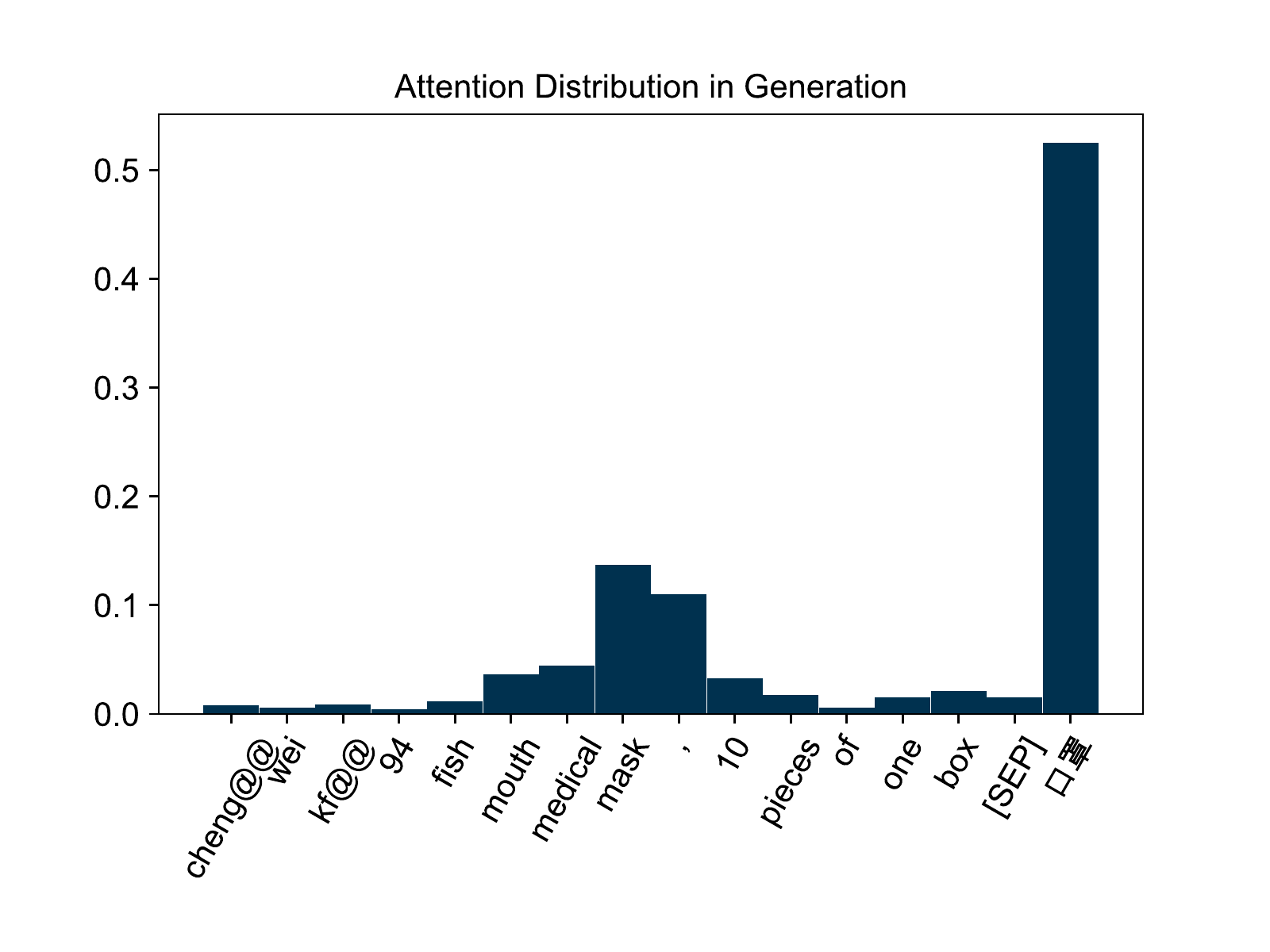}
    \caption{The attention distribution when predicting the token  ``口罩''  (``mask'' in English). The model shows a strong attention preference to the prompt words. }
    \label{fig:attention}
\end{figure}
\input{tables/cases.tex}
\paragraph{\soft} We verify how much influence the visual signal imposes upon the model. 
Inspired by \citet{WuKBLK20}, we quantify the modality contribution via the L2-norm ratio 
($\Lambda H^{\mathrm{fused}}$ for vision over 
$ H^{\mathrm{text}}$ for text, in Eq.~\ref{main}). 
We visualize the whole training process along with BLEU as a reference in Figure~\ref{fig:ratio-bleu}.
\citet{WuKBLK20} criticize that previous studies do not utilize visual signal, for the final ratio converge to zero.
Our method shows a different characteristic: as the BLEU becomes stable, the ratio of visual signal and textual signal still remains at around $0.15$, showing the effectiveness of the visual modality.

\paragraph{\hard} We also look into the influence caused by the prompts. We sample an ambiguous sentence: ``\textit{chengwei kf94 fish mouth medical mask, 10 pieces of one box}''. The keyword ``mask'' can be translated into ``口罩'' (``\textit{face mask}'' in English) or ``面膜'' (``\textit{facial mask}'' in English) without any context. We visualize the attention distribution when our \hard model is translating ``mask'' in Figure~\ref{fig:attention}. We can see that the a high attention is allocated to the caption prompt. Therefore, our method correctly translates the word.
We also visualize the detailed attention heatmaps for source, prompts and generated sentences in Appendix.

\subsection{Qualitative Case Study}

We also compare several cases from \data testsets to discuss how multimodal information and external bilingual data help the translation performance. Meanwhile, we regard the case study as a spot check for the multimodal translation testset itself. 
We here choose plain Transformer, our methods trained on triplet only and all data, as well as human reference for comparison.

Table~\ref{tab:case_study} presents the qualitative cases and major conclusions are as follows: 1) Visual information plays a vital role in disambiguating polysemous words or vague descriptions. 2) Non-triple data improves translation accuracy, particularly in translating jargons and enhancing fluency in the general lack of multimodal data. 3) Our test set is representative in real-world seniors as it includes product titles that are confusing and require image, in contrast to previous case studies on Multi30k where researchers artificially mask key words~\cite{CaglayanMSB19, WuKBLK20, WangX21, LiLZZXMZ22}.

%% file: tables/cases.tex
\begin{table*}[h]
	\centering
	\resizebox{.83\textwidth}{!}
	{%
	\begin{tabular}{cll@{}}
	\toprule
	\MR{8}{*} {\includegraphics[height=2.1cm]{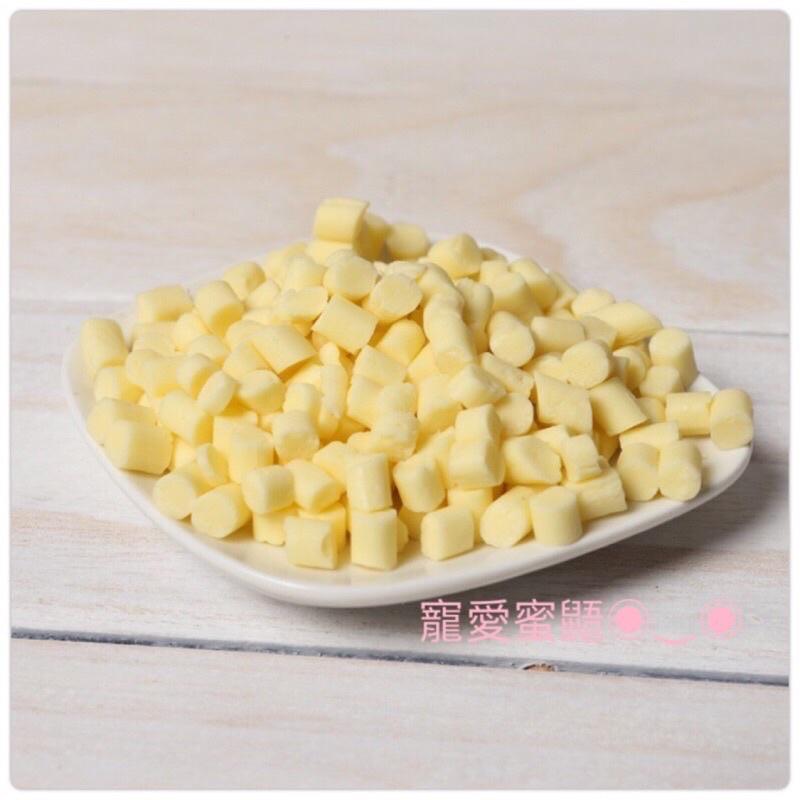}} 
	& \T{source:} & ready stock , cheese grains , pets only \\
	& \T{human:} &  现货\ 商品\ 奶酪\ 粒\ (宠物\ 专用） \\
	& \T{Plain:} &  现货\ 起司\ \sout{谷物}\ 宠物\  \textbf{专用} \\
	& \phantom{\T{ Plain:}} &  ready stock cheese \sout{cereal grains} pet only \\
	& \T{Ours (Triplet-only):}  & 现货\  奶酪\  \underline{颗粒}\ 宠物 \ \sout{仅限}  \\
	& \phantom{\T{\method:}} &  ready stock cheese \underline{granular} pets \sout{only for} \\
	& \T{Ours (All-data):} &  现货 \ 奶酪 \ \textbf{粒} \ 宠物 \ 专用 \\
	& \phantom{\T{\methodext:}} &  ready stock  cheese \textbf{grains} for pets only \\

    \midrule
	\MR{8}{*} {\includegraphics[height=2.1cm]{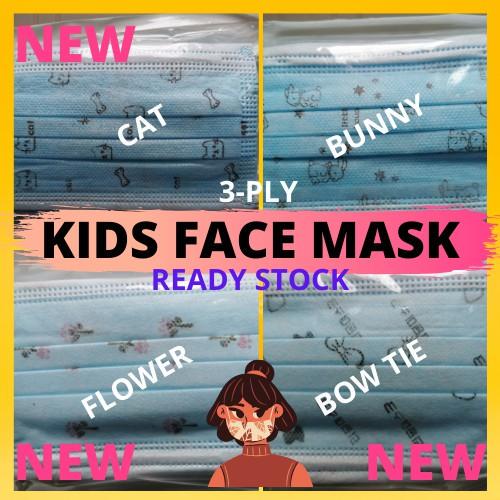}} 
	& \T{source:} & ready stock kids medical surgical face mask 3-ply 20pcs \\
	& \T{human:} &  现货\ 儿童\ 医疗\ 手术\ 口罩\ 3\ 层\ 20\ 个 \\
	& \T{Plain:} &  现货\ 儿童\ 医用\ \sout{面具}\ 3\sout{-ply}\ 20\sout{pcs}\ \\
	& \phantom{\T{ Plain:}} &  ready stock kids medical \sout{(opera) mask} 3-\sout{ply} 20\sout{pcs}\\
	& \T{Ours (Triplet-only):}  & 现货\ 儿童\ 医用\ \textbf{口罩}\ 3\sout{ply}\ 20\sout{pcs}\ \\
	& \phantom{\T{\method:}} &  ready stock kids medical \textbf{mask} 3-\sout{ply} 20\sout{pcs}  \\
	& \T{Ours (All-data):} &  现货\ 儿童\ 医用\ 外科\ \textbf{口罩}\ 3\ \textbf{层}\ 20\ \textbf{片}\ \\
	& \phantom{\T{\methodext:}} &  ready stock kids medical surgical \textbf{mask} 3-\textbf{ply} 20\textbf{pcs} \\
	
	\bottomrule
	\end{tabular}
	}
	\caption{Qualitative examples from two complex scenarios. Plain and Ours (Triplet-only) respectively indicate plain Transformer and \method trained on triple data only, and  Ours (All-data) indicate \method trained on all data.  The \sout{strikethrough},  \underline{underline} and \textbf{bold}  indicates inappropriate, reluctant and excellent choices respectively. 
	The more detailed comments on the translation are in Appendix. }
	\label{tab:case_study}
\end{table*}

%% file: 080conclusion.tex
This paper devises a new framework \method for multimodal machine translation and introduces two approaches to utilize image information. The new methods are effective and highly interpretable. 
Considering the fact that current multimodal benchmarks are limited and biased, we introduce a new dataset \data of the e-commercial domain. To better validate the multimodal translation systems, the testset is carefully selected as the image are crucial for translation accuracy. 
Experimental results and comprehensive analysis show that \method makes a strong baseline and \data can be a promising benchmark for further research.

%% file: 085limitation.tex
First, there are studies~\cite{WuKBLK20} claiming visual information only serves as regularization. In our ablation study, we find the adversarial setting of \soft approach outperforms the plain Transformer. Combined with observations from previous studies, we suggest that fusion-based architectures may apply some images information as regularization terms, yet the further quantitative analysis is needed to confirm this phenomenon.

Second, though our testset is carefully selected to ensure the textual ambiguity without image data, we encounter difficulties in designing a suitable metric for quantifying the degree to which the models are able to resolve the ambiguity.
Specifically, we find that conventional metrics, such as word-level entity translation accuracy, exhibit significant fluctuations and do not effectively quantify the extent to which the model effectively resolves ambiguity.
We discuss this metric in more details in the Appendix, and offer a glossary of ambiguous words used in the test set. We acknowledge that the evaluation of multimodal ambiguity remains an open problem and an area for future research.

In addition, there are some details regarding the dataset that we need to clarify: the dataset is collected after COVID-19, so some commodities will be associated with the pandemic. We collect data by category in order to cover various products to reduce the impact of the epidemic on product types.

%% file: 090appendix.tex
\subsection{Ethic Consideration About Data Annotators}
We hire 20 professional translators in a crowd-source platform and pay them according to the market wage and work within 8 hours a day.
All translators are native Chinese and have graduated with an English major. 
The ethics review is done while in data acceptance stage.

\subsection{Data Copyright}

In our study, we present a new dataset of public e-commercial products from Shoppee and TikTok Shop. To address copyright concerns, we provide a detailed description of how we collect the data and ensure that our usage complies with all relevant policies and guidelines.

For the Shoppee dataset, we obtain the data from their Open Platform API\footnote{https://open.shopee.com/documents}. We carefully review their Data Protection Policy \footnote{https://open.shopee.com/developer-guide/32} and Privacy Policy guidelines \footnote{https://careers.shopee.tw/privacy-policy}, which provide clear instructions for using data through the Shopee Open Platform. We strictly follow their requirements and limitations, ensuring that we did not access any personal data and that we only use open information provided by the API. We also adhere to their robot guidelines \footnote{https://shopee.tw/robots.txt}, avoiding full-site scraping.

For the TikTok Shop dataset, we access the data using robots, as scraping is allowed according to their robots.txt file \footnote{https://shop.tiktok.com/robots.txt}.  We also review TikTok Shop Privacy Policy and TikTok for Business Privacy Policy \footnote{https://tiktokfor.business/privacy-policy/} to ensure that we only collect data from merchants under their policy.

It is important to note that all data we publish is publicly available on the Internet and only pertains to public e-commercial products. We do not access or publish any user information, and we take all necessary steps to respect the intellectual property and privacy rights of the original authors and corresponding websites. If any authors or publishers express a desire for their documents not to be included in our dataset, we will promptly remove that portion from the dataset. Additionally, we certify that our use of any part of the datasets is limited to non-infringing or fair use under copyright law. Finally, we affirm that we will never violate anyone's rights of privacy, act in any way that might give rise to civil or criminal liability, collect or store personal data about any author, infringe any copyright, trademark, patent, or other proprietary rights of any person.

\subsection{Results on Fashion-MMT}

We list the testset performance on Fashion-MMT in Table~\ref{tab:fmmt}. 

\input{tables/fashionMMT.tex}

\input{tables/bleurt.tex}
Fashion-MMT is divided into two subset according to the source of the Chinese translation: ``Large'' subset for the machine-translated part and ``Clean'' subset for the manually annotated part. As its authors also found the Large subset is noisier and different from the human annotated data, our experiments focused on the Clean subset with Fashion-MMT(\ie Fashion-MMT(c)).

We compare the model performance on training on Triplet Only and adding Parallel Text settings. As the original dataset does not provide a parallel corpus without pictures, we used Parallel Text from \data for our experiments.

Note that the UPOC$^2$ model relies on three sub-methods, namely MTLM, ISM, and ATTP. The ATTP requires the use of commodity attributes, whereas our model does not use such information. Hence, we also list results of UPOC$^2$ without ATTP in the table.

The results show that our model rivals UPOC$^2$ on triplet only settings. And by using parallel text, ours gain further improvement, even if the parallel text does not match the domain of the original data. The results demonstrate the potential of our training strategy over multiple domains.

\subsection{Evaluation with various metrics}

Recent studies have indicated that the sole reliance on BLEU as an evaluation metric may be biased~\citep{MarieFR20}. We hence evaluate models with machine learning-based metric BLEURT~\citep{SellamDP20} and list the results in Table~\ref{tab:bleurt}\footnote{We use BLEURT-20 model from \url{https://github.com/google-research/bleurt}}. 

Previous multimodal works often set entity nouns in the original sentence into [mask] to quantify model's ability for translating masked items with images~\citep{WangX21,LiLZZXMZ22,FangF22}. While the experiment can measure the effectiveness of multimodal information, text with [mask] is not natural and the setting makes less sense in the real world. 
Inspired by their settings, we have developed a set of commonly used English-Chinese translation ambiguities by mining frequently used product entity and manual annotating. We have defined an word-level accuracy metric based on those potential ambiguous words in Table~\ref{tab:amb}: if a certain English word appears in the original sentence, we require that the model's translation result in the target language must be consistent with the human reference's corresponding entity translation in order to be considered a correct translation, and thus calculate the word-level accuracy.

The results of BLEURT generally align with BLEU, indicating the effectiveness of \method. However, an exception occurs in the Google Translate system, whose score are highest among all systems. We attribute this deviation to the use of back-translated pseudo corpus in the pre-training of the BLEURT model. 

Multimodal models consistently perform better than plain transformer models in word-level accuarcy. Additionally, Google Translate obtains the lowest scores in word-level accuracy, indicating that BLEURT may not  distinguish ambiguous words in multimodal scenarios. However, the  difference between multimodal ones is not significant. We attribute it to  the difficulty in quantifying the semantic differences between synonyms, as we will demonstrate in our case study details. Furthermore, given the significant human effort required for mining and annotating ambiguous word list while it is highly domain-specifc to the test set, we suggest that the development of new metrics for evaluating multimodal translation ambiguity shall be a valuable topic of future research.

\subsection{Translation Details of Case Study}

Here we give some detailed explanations about the translation of case study translations:

In the first case, the Plain Transformer fail to recognize whether the word ``grains'' means cereal crop (谷物) or the cheese of grain sizes(奶酪粒).
Triplet-Only \method translate ``grains'' into 颗粒, which is acceptable, but the word not commonly used to describe food in Chinese, yet the model does not translate "only" grammatically properly. 

In the second case, Plain Transformer translates ``mask'' to 面具, which is more commonly used to refer opera mask in Chinese. Both Plain Transformer and Triplet-Only \method fail to understand ``pcs''(件、个、片) and ``ply''(层), and directly copy them to targets. The two methods also fail to translate ``surgical''(手术、外科) correctly as it is a rare word in Triplet only settings.

In comparison, the translation of \method is more consistent with the images, and more appropriate in terms of grammar and wording.

\subsection{Attention Visualization}

\begin{figure*}[t]
	\centering
	\includegraphics[width=0.9\linewidth]{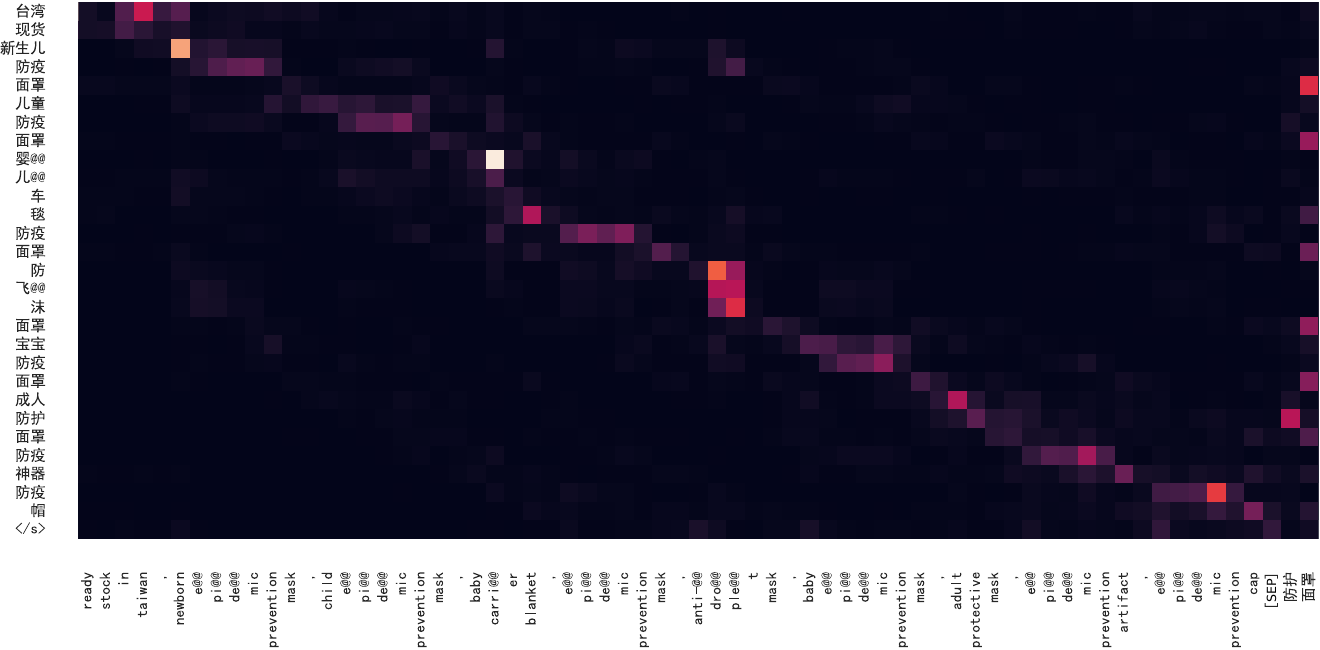}
	\caption{Attention heat map between hypo sentence and the source with caption prompts}
	\label{fig:heatmap2}
\end{figure*}

We visualize one the attention heatmap case of \hard in Figure~\ref{fig:heatmap2} and Figure~\ref{fig:heatmap2}. 

Figure~\ref{fig:heatmap2} shows the attention alignment of original source (y-axis) and the prompted source (x-axis)  in text encoder.
Figure~\ref{fig:heatmap2} shows the generated sentence (y-axis) and the prompted source (x-axis)  in text decoder.
\input{tables/extra.tex}
From the heat map we know that the prompt attends to the most relevant ambiguous words and supports the model translation, both when encoding the source sentence and decoding the infernece.
Specifically in our case, ``口罩''(face mask) in prompts has high attention with all ``masks'' occurrence on the source side, and has  high attention with all ``口罩'' generation in decoder side. In contrast, the word ``防护''(protective)  less prominent in the attention heatmap as it is less ambiguous.


\subsection{Details on Data Selection and Mixing}

As discussed in Section~\ref{sec:setups}, we resort to upsampling the e-commercial triplet data due to the significant disparity in the quantity of data across various domains. As previously proposed by ~\citet{WangN19} and ~\citet{abs-1907-05019}, we utilize a temperature-based sampling method, where the i-th data split is assigned a sampling weight proportional to $D_i^{\frac{1}{T}}$, where $D_i$ denotes the number of sentences in the i-th data split, and $T$ is the temperature hyper-parameter. In our implementation, to guarantee the completeness and homogeneity of data across each training iteration, we directly upsample the triplet data or monolingual captions, and subsequently, shuffle them randomly with parallel text to construct the training dataset. The upsampling rate for the triplet data is rounded to 15 and the upsampling rate for the parallel text is rounded to 4, resulting in an actual sampling temperature of 5.11~.
\input{tables/amb.tex}

\section{Model Performance with Excessive Data}

Based on data distribution and scaling laws, we sample 750k parallel text and 103k monolingual captions as non-triple data to validate our methods. To further explore the potential of models with excessive non-triple data, we attempt to increase the data scale of the parallel text corpus to 5M, which are also sampled from CCAlign corpus. We list the results in Table~\ref{tab:extra}. 
However, we find that excessive parallel text does not further promote model performance on current test sets.
We suggest that the lack of improvement in performance may be due to the difference in text domain between the general domain and the e-commerce domain. 
As we will release the parallel text corpus we used in our experiments, in addition to conducting fair comparisons based on our data, we also encourage future researchers to use more unconstrained external data and techniques to continue to improve performance.

\begin{figure*}[t]
	\centering
	\includegraphics[width=0.9\linewidth]{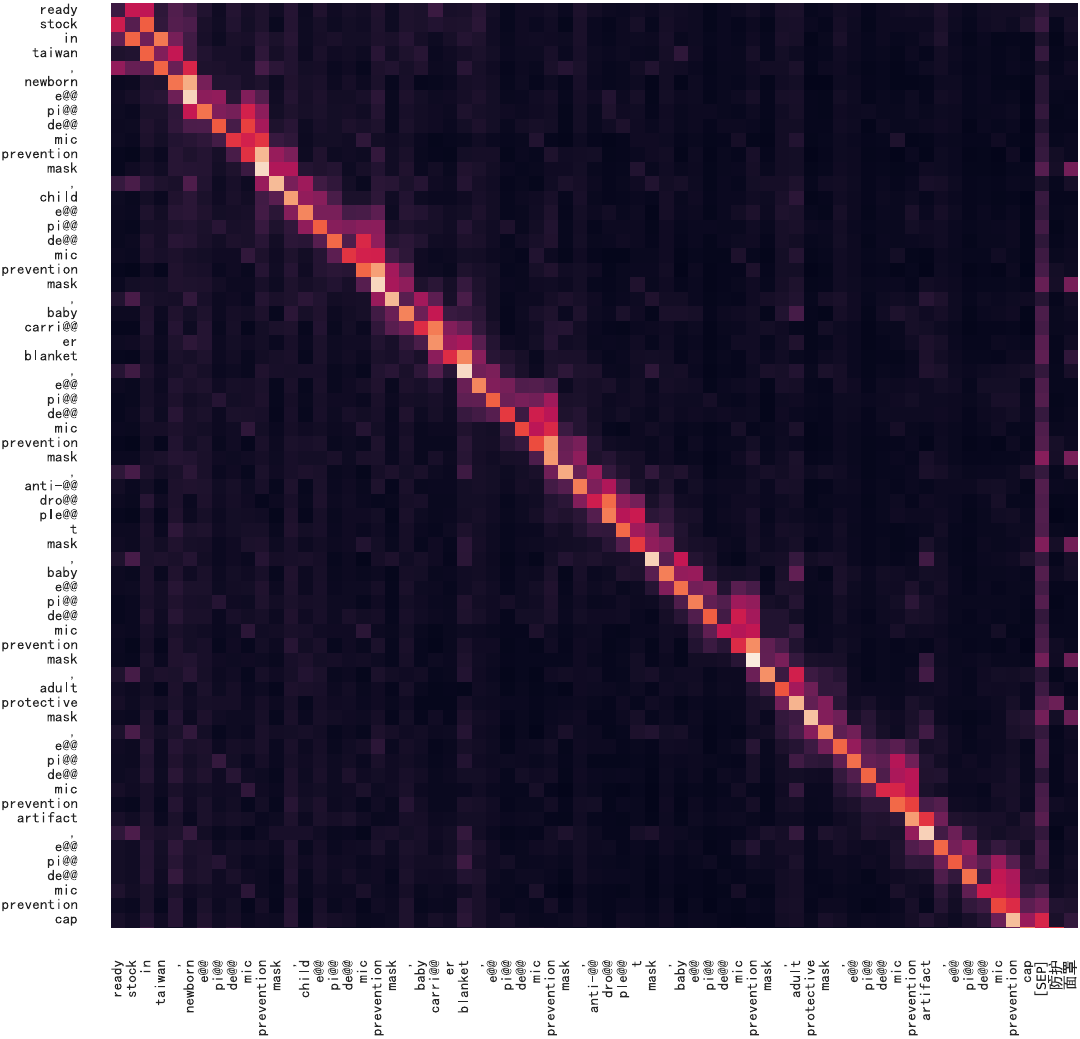}
	\caption{Attention heat map between source sentence and the source with caption prompts}
	\label{fig:heatmap}
\end{figure*}

%% file: tables/fashionMMT.tex
\begin{table}[h!]
\scriptsize
\centering
\begin{tabular}{c|l|cc}
\hline
\multicolumn{1}{l|}{}  & FashionMMT（C）                           & Triplet Only & + Parallel Text \\ \hline
\multicolumn{1}{l|}{}  & Transformer                             & 40.12        & /               \\ \hline
\multirow{2}{*}{UPOC$^{2}$} & MTLM+ISM                                & 41.38        & /               \\
                      & MTLM+ISM+ATTP                           & 41.93        & /               \\ \hline
\multirow{3}{*}{Ours} & \textsc{Fusion}        & 41.19        & 42.38           \\
                      & \textsc{Prompt}        & 40.97        & 42.02           \\
                      & \textsc{Fusion+Prompt} & 41.38        & 42.33           \\ \hline 
\end{tabular}

\caption{Results on Fashion-MMT(C) testset.  }
\label{tab:fmmt}
\end{table}

%% file: tables/bleurt.tex
\begin{table*}[]

\begin{tabular}{l|l|ccc|ccc}
\Xhline{3\arrayrulewidth}
\hline
\multirow{2}{*}{ID} &  Metric      & \multicolumn{3}{c|}{BLEURT}                         & \multicolumn{3}{c}{Accuracy}              \\ \cline{2-8} 
                    &  Training Data             & Triplet Only    & + PT            & + PT + MC       & Triplet Only & + PT           & + PT + MC \\ \hline
1                   & Plain Transformer   & 0.5424          & 0.5559          & 0.5662          & 0.765        & 0.754          & 0.761     \\
2                   & Selective Attention & 0.5619          & /               & /               & 0.782        & /              & /         \\
3                   & UPOC$^2$            & 0.4855          & /               & 0.5788          & 0.792        & /              & 0.798     \\ \hline
4                   & UVR-NMT             & 0.5299          & 0.5866          & /               & 0.795        & 0.791          & /         \\
5                   & Phrase Retrieval    & /               & /               & /               & /            & /              & /         \\ \hline
6                   & \soft                & \textbf{0.5760} & 0.5782          & 0.5923          & 0.771        & 0.778          & 0.812     \\
7                   & \hard                & 0.5600          & 0.5772          & 0.5980          & 0.792        & 0.775          & 0.792     \\
8                   & \textsc{Fusion+Prompt}                 & 0.5647          & \textbf{0.5917} & \textbf{0.6018} & 0.809        & \textbf{0.791} & 0.799     \\ \hline
                    & Google Translate    & \multicolumn{3}{c|}{\textbf{0.6108}}                & \multicolumn{3}{c}{0.741}                 \\ \hline
                    \Xhline{3\arrayrulewidth}
\end{tabular}

\caption{Results of \method and baselines on \data evaluated by BLEURT and word-level accuarcy of ambiguous words. }
\label{tab:bleurt}%

\end{table*}

%% file: tables/extra.tex
\begin{table}[]
\centering
\begin{adjustbox}{max width=0.45\textwidth}
\begin{tabular}{l|cccc}
\hline
Metric            & \multicolumn{2}{c}{BLEU} & \multicolumn{2}{c}{BLEURT} \\ \hline
Data              & + PT    & + Excessive PT & + PT     & + Excessive PT  \\ \hline
Plain Transformer & 40.66 & 40.86        & 0.5559   & 0.5406          \\
\soft              & 44.22 & 43.91        & 0.5782   & 0.5877          \\
\hard              & 43.35 & 43.61        & 0.5772   & 0.5760          \\
\textsc{Fusion+Prompt}              & 45.20 & 44.87        & 0.5917   & 0.6025          \\ \hline
\end{tabular}
\end{adjustbox}

\caption{Results of \method and plain Transformer on \data  with parallel text and excessive parallel text~(5M).  
}
\label{tab:extra}

\end{table}

%% file: tables/amb.tex
\begin{table*}[]
\scriptsize 
\centering
\begin{tabular}{ll|ll}
\hline
English Word & Chinese Potential Translations & English Word   & Chinese Potential Translations \\ \hline
mask         & 面膜,口罩,面罩,面具,遮垫                 & tape           & 胶带,胶布,带子,磁带,薄胶带                \\
bow          & 琴弓,弓子,弯弓                       & bar            & 吧台,酒吧,棒杆                       \\
top          & 上衣,上装,女上装,机顶                   & basin          & 盆子,盆器,盆,地盆,盆池                  \\
set          & 套装,把套,撮子,套盒,组套                 & sheet          & 被单,棚布,薄板,薄片,片材                 \\
clip         & 卡子,提盘夹,提盘夹子,夹片,取夹              & film           & 贴膜,薄膜,胶片,胶卷,软片                 \\
nail         & 钉子,铁钉,扒钉,指甲,钉钉子                & eyeliner       & 眼线笔,眼线液,眼线,眼线膏                 \\
iron         & 铁,铁艺,电熨斗,熨斗,烫斗                 & shell          & 车壳,被壳,贝壳,外壳,壳壳                 \\
rubber       & 胶皮,橡皮                          & chip           & 芯片,筹码                          \\
brush        & 刷子,毛笔,毛刷,板刷,锅刷                 & plug           & 插头,塞子,胶塞,堵头,地塞                 \\
oil          & 机油,油,油脂,油液                     & napkin         & 餐巾纸,餐巾                         \\
canvas       & 餐布,油画布,画布,帆布                   & grease         & 润滑脂,打油器                        \\
ring         & 戒指,指环,圆环,圈环,响铃                 & pipe           & 管子,烟斗,管材,皮管,排管                 \\
pad          & 护垫,盘垫,踏垫,垫块,贴垫                 & charcoal       & 木炭,炭笔,炭,引火炭,炉炭                 \\
wipes        & 湿巾,抹手布,擦地湿巾,擦碗巾,擦奶巾            & blade          & 铲刀,刀片,叶片,刀锋,遮板                 \\
face mask    & 焕颜面膜,护脸面罩,遮脸面罩,脸罩,脸部面膜,口罩      & bucket         & 水桶,面桶,扒斗,漂桶,簸箩                 \\
powder       & 粉饼,散粉,粉掌,修容粉饼,粉剂               & lift           & 升降机,升降梯,举升机,举升器,起重器            \\
tie          & 扎带,领带                          & crane          & 吊车,起重机,吊机,起重吊机,仙鹤              \\
desktop      & 桌面,台式机                         & football       & 足球,橄榄球                         \\
jack         & 千斤顶,插孔                         & frame          & 画框,车架,框架,包架,裱画框                \\
collar       & 项圈,颈圈,套环,领夹                    & plum           & 话梅,李子                          \\
cement       & 胶泥,水泥                          & slide          & 滑轨,滑梯,滑滑梯,滑道,幻灯片               \\
tank         & 坦克,料槽,坦克车                      & keyboard       & 键盘,钥匙板,小键盘                     \\
hood         & 头罩,遮光罩,机罩,风帽,引擎盖               & bass           & 鲈鱼,贝斯                        \\
gum          & 牙胶,树胶,口香糖                          & makeup remover & 卸妆水,卸妆膏,卸妆液,卸妆乳,卸妆棉棒           \\
screen       & 屏风,纱窗,滤网,丝网,筛网                 & counter        & 计数器,柜台                         \\
bell         & 铃铛,车铃,吊钟,吊铃                    & separator      & 隔板,分离器,分液器,隔片,分离机              \\ \hline
\end{tabular}

\caption{Potential ambiguous product entities in English-Chinese translations. The Chinese translations are separated by commas and have different meanings in the alternative translations. Words are sorted by the frequency in e-commercial English corpus}
\label{tab:amb}

\end{table*}